\documentclass[runningheads]{llncs}
\usepackage{graphicx}
\usepackage{multirow}
\usepackage{subcaption}
\begin{document}
\title{Explaining Clustering of Ecological Momentary Assessment Data
Through Temporal and Feature Attention}
\titlerunning{Explaining EMA Clustering Through Attention}
\author{Mandani Ntekouli\inst{1} \and
Gerasimos Spanakis\inst{1} \and
Lourens Waldorp\inst{2} \and Anne Roefs\inst{3}}
\authorrunning{M.Ntekouli et al.}
\institute{Department of Advanced Computing Sciences, Maastricht University, Maastricht, The Netherlands \email{\{m.ntekouli, jerry.spanakis\}@maastrichtuniversity.nl} \and
Department of Psychological Methods, University of Amsterdam, Amsterdam, The Netherlands \email{L.J.Waldorp@uva.nl} \and
Faculty of Psychology and Neuroscience, Maastricht University, Maastricht, The Netherlands\\
\email{a.roefs@maastrichtuniversity.nl}}
\maketitle              % typeset the header of the contribution
\begin{abstract}
In the field of psychopathology, Ecological Momentary Assessment (EMA) studies offer rich individual data on psychopathology-relevant variables (e.g., affect, behavior, etc) in real-time. EMA data is collected dynamically, represented as complex multivariate time series (MTS). Such information is crucial for a better understanding of mental disorders at the individual- and group-level. More specifically, clustering individuals in EMA data facilitates uncovering and studying the commonalities as well as variations of groups in the population. Nevertheless, since clustering is an unsupervised task and true EMA grouping is not commonly available, the evaluation of clustering is quite challenging. An important aspect of evaluation is clustering explainability. Thus, this paper proposes an attention-based interpretable framework to identify the important time-points and variables that play primary roles in distinguishing between clusters. A key part of this study is to examine ways to analyze, summarize, and interpret the attention weights as well as evaluate the patterns underlying the important segments of the data that differentiate across clusters. To evaluate the proposed approach, an EMA dataset of 187 individuals grouped in 3 clusters is used for analyzing the derived attention-based importance attributes. More specifically, this analysis provides the distinct characteristics at the cluster-, feature- and individual level. Such clustering explanations could be beneficial for generalizing existing concepts of mental disorders, discovering new insights, and even enhancing our knowledge at an individual level. 

\end{abstract}
\section{Introduction}
For quite some years now, Ecological Momentary Assessment (EMA) \cite{becker2016predict,torous2018smartphones} has been a popular method in psychology research to collect data in daily life. During an EMA study, individual real-time data is collected over time, capturing various psychopathology-related behaviors, experiences, and symptoms in their natural environments \cite{fried2017moving}. For each individual, EMA data is then organized as a multivariate time series (MTS). Such a rich structure sets new pathways to employing advanced analytical techniques, enhancing our understanding of the complex patterns in psychopathology.

According to psychopathology theory, personalized models have been traditionally applied. A popular model for EMA data is the vector Autoregressive (VAR) model, where one or more time lags are used to predict variables at subsequent time-points. Because of heterogeneity between individuals, these models focusing on each individual separately allow for interventions and treatments that are tailored to their unique patterns of symptoms~\cite{roefs2022new}. Although, personalized models can reveal quite distinct patterns, they may also reveal commonalities. Such commonalities can be quite valuable, giving insights into more general mechanisms that are valid for particular subgroups. It is, however, quite difficult to ascertain if there are such subgroups for which commonalities hold and also how to discover them. 

In an EMA study, collecting data for multiple individuals at the same time could facilitate the understanding of mental disorders by studying the commonalities as well as variations present in the sample. An interesting approach for broadening EMA analysis involves enhancing models with data from more than one individual~\cite{ntekouli2022using}. More specifically, group-based models, relying on individuals exhibiting similar temporal patterns of symptomatology or behavior, can offer potentially valuable insights for the whole sample~\cite{ntekouli2023modelbased}. Thus, uncovering homogeneous groups of people is of great importance. Such groups of individuals can be uncovered by clustering applied to EMA data~\cite{ntekouli2023evaluating}. Nevertheless, since clustering is an unsupervised task and true EMA grouping is not commonly available, the evaluation of the derived clustering labels is quite challenging. Because of the inherent variability of EMA data as well as the complexity of the real-world MTS, it is expected that clustering algorithms could produce quite different results leading to different groups. Thus, a critical point to be addressed is evaluating the clustering results. 

Beyond the well-applied distance-based criteria capturing the quality of clusters, clustering interpretability is another important aspect of evaluation. Interpretability ensures that the patterns and common characteristics of the groups identified through clustering can be understood, explained and validated in the context of mental disorders. Thus, uncovering the meaningful patterns of each cluster within EMA data could provide insight into intra-individual psychopathological processes, their temporal patterns, and their inter-relationships among theoretically similar subtypes of disorders. 

\begin{figure}[t]
    \centering
    \includegraphics[width=0.9\textwidth]{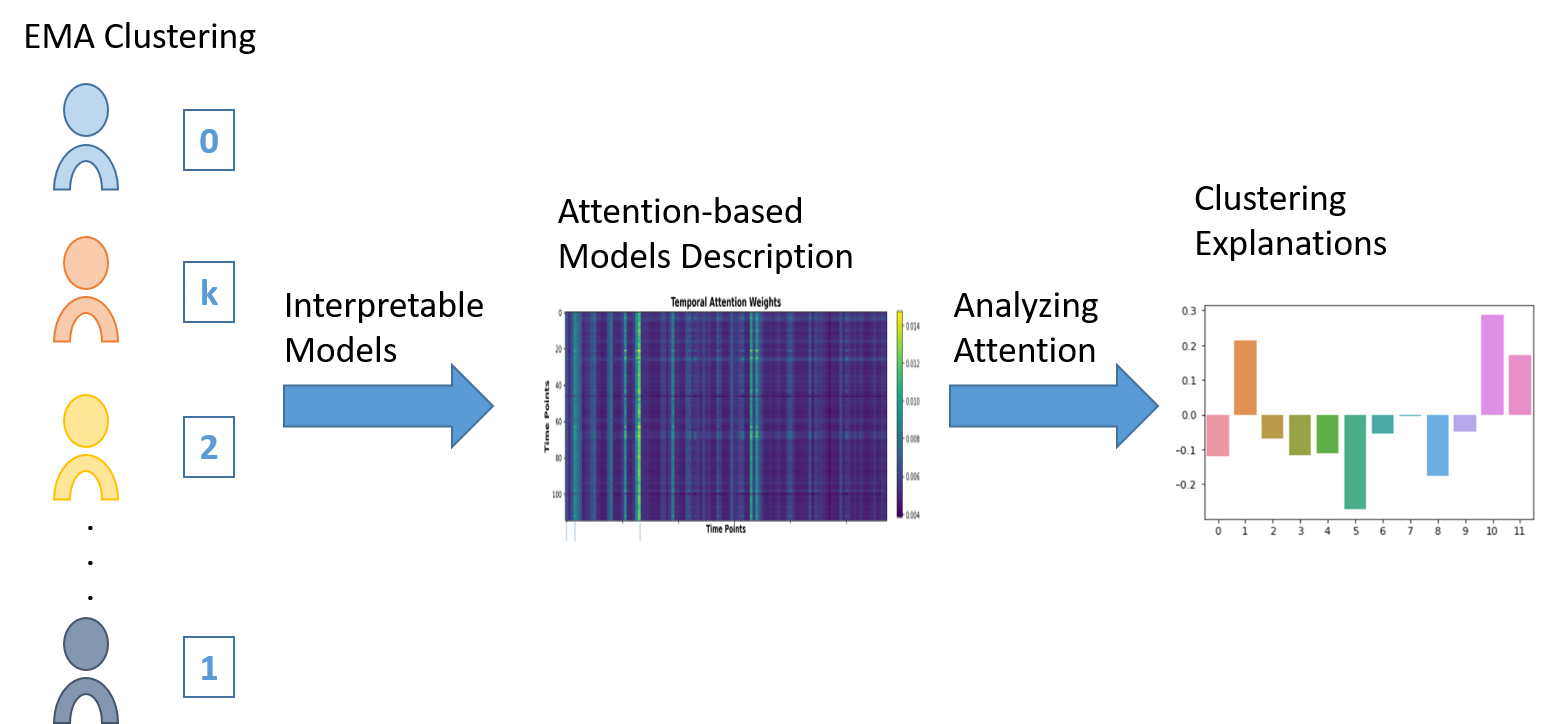}
    \caption{An overview of our methodological approach focusing on applying attention-based interpretable models to provide clustering explanations.}
    \label{fig:first}
\end{figure}

In this paper, to address the need for explanations on clustering results, the proposed methodology focuses on investigating interpretable deep-learning mechanisms capable of handling MTS data, particularly within the domain of psychopathology. Our approach utilizes advanced interpretative deep-learning models, specifically attention-based mechanisms \cite{vaswani2017attention}, to understand the complexities inherent in MTS data without relying on prior data transformations. Therefore, this approach ensures that our interpretations rely on the actual data dynamics rather than other transformations, providing a clearer and more accurate comprehension of MTS in psychopathology. As depicted in Figure~\ref{fig:first}, our methodology employs a multi-aspect framework that integrates both temporal and feature-level attention. Therefore, it is designed to provide explanations by identifying the important time-points and variables that play primary roles in the domain of psychopathology \cite{vskrlj2020feature,hsieh2021explainable}. While the proposed framework for extracting temporal and feature-level attention is fully described, a significant part of this paper is to examine ways to analyze, summarize, and interpret the attention weights as well as validate the patterns underlying the important segments of the data that differentiate between clusters. More specifically, this includes the significant and distinct characteristics in cluster-, feature- and individual-level and their evaluation. Thus, such clustering explanations could be proven beneficial for generalizing the existing concepts, uncovering new insights into psychopathology and network theory, and even enhancing our knowledge at an individual level. Furthermore, central to our framework is its independence from any specific clustering algorithms. This independence introduces a new theoretical perspective on evaluating the robustness of an examined clustering result, allowing for a more objective comparison and assessment of clustering algorithms. Overall, attention-derived interpretability, beyond contributing to a better understanding of the underlying data structures in psychopathology, could be theoretically used to benchmark clustering effectiveness.

\section{Related Work}
The field of time-series (TS) clustering research has attracted significant attention, with a focus on clustering explanations being introduced the recent years. First, related work on time-series clustering explanations is presented. This involves extracting meaningful representation for clusters' descriptions as well as the classical explanation methods of the field of Explainable Artificial intelligence (XAI). 

\subsection{Clusters' Descriptive Representation}
In the context of understanding clustering results, the representation of clusters is crucial in unveiling meaningful insights into underlying constructs (connections in variables) and further facilitating decision-making. This is usually achieved by examining and visualizing all elements of a cluster. For instance, in temporal data, distinct trend lines can be observed by overlying the time series data belonging to each cluster. Summary statistics, such as mean values and variance, can be also calculated to describe each cluster. However, the difficulty of distinguishing between them increases significantly when using many clusters or high-volume datasets with multiple variables.

\subsection{Explanations on TS Clustering}
Beyond statistically analyzing and exploring the structures of each cluster, the majority of the work focuses on extracting the factors (e.g. important variables) that influence the cluster assignments by applying interpretable models. The methodologies employed for explaining time-series clustering vary considerably, despite the low number of publication outputs. 

Following the approach of transforming the data to different time-series representations, the work of ~\cite{time2feat} applies clustering on various interpretable extracted features, including intra-signal (or within feature) and inter-signal (across different features) characteristics, such as variance and correlation, respectively. After data pre-processing and dimensionality reduction techniques, a selection of features is retained for clustering, showing the high importance of inter-signal features as always preserved in clustering. The number of features is then used as a measure of interpretability, meaning that fewer features facilitate clusters’ comprehension. 

Moreover, other work, like \cite{local,xclusters} was inspired by the general trend of training and explaining classification models to predict the derived cluster labels. On the one hand, according to \cite{local}, local interpretability methods, such as Local Interpretable Model-Agnostic Explanations (LIME), SHapley Additive exPlanations (SHAP) as well as Gradient-weighted Class Activation Mapping, are used with different classification models (e.g. XGBoost) and trained again on statistically-extracted temporal features. Examples of such features were auto-correlation and median difference. 
On the other hand, the work of ~\cite{xclusters} proposed a different approach to holistically perform clustering and provide explanations through training a decision tree. In practice, to achieve this, two different data sources were used, time-series and static or baseline data (such as demographic data), for clustering and interpretation, respectively. After applying clustering on TS, the cluster labels accompanied by the associated static data were used to train an interpretable decision tree model. Therefore, it aims to optimize both objectives, clustering and interpretation, at the same time.

According to the studies above, it becomes evident that existing methodologies for providing explanations rely on the applied transformations of the original data. This means that explanations also refer to the transformed feature space rather than the raw features raising some questions regarding the actual interpretability and transparency of the explanations. Therefore, beyond the limited work on the topic, there is a need for further exploration of the level of the original data.

\section{Review on Challenges of Explaining MTS Data}
Due to the inherent challenges of the topic and limited available published work, the current paper was also inspired by the literature focusing on using interpretable classification models for explaining an output vector, where the classes are the distinct cluster labels. In our setting, that output can be the clustering-derived result. Thus, an additional review of the challenges of applying interpretable classification models on MTS is necessary.

\subsection{Clustering Explanations}
Although it is acknowledged that the interpretability of clustering is of great importance in uncovering meaningful insights about data structures, limited work has been conducted in parallel with developing clustering algorithms. All the well-known clustering methods were designed to group data, mainly considering various objective functions and dealing with different data types, but without taking into account any interpretability aspects of clustering. Therefore, when clustering results need to be explained, post-processing steps are commonly used, involving an additional classification model trying to predict and explain the clustering labels.  
For instance, in ~\cite{x-kmeans,x-kmeans2}, interpretable threshold decision trees having $k$ leaf nodes are applied to explain the labels of k-means or k-medoids. Therefore, in most cases, the problem of clustering interpretability can be formulated as classification interpretability. This is a quite more studied issue, with a goal to explain the cluster labels as classification targets. 

Regarding classification explainability, beyond using inherently interpretable models, such as linear or decision tree models, post-hoc explainability methods are also commonly explored. These fall into two main categories, model-specific and model-agnostic methods. Model-specific techniques correspond to particular groups of models, such as extracting feature importance from tree-based models and tree ensembles or utilizing layer-specific integrated gradients for deep-learning models \cite{integrated_grad}.
On the other hand, model-agnostic explanation methods are applicable on top of various models regardless of their architecture. More specifically, these include the widely-applied LIME focusing on generating local and instance-specific explanations, as developed by \cite{lime}, and SHAP for feature-based explanations developed by \cite{shap}. 

However, when dealing with MTS clustering, challenges arise in two aspects of such formulation. First, the aforementioned XAI techniques of supervised learning models typically focus on images, text, and tabular data, limiting their application to time-series data. Second, it is not straightforward how to input time-series data into the classical classification models, without discarding their temporal nature. Therefore, adjustments should be made in both parts to better explain the dynamics of MTS data.

\subsubsection{Adapting XAI methods to MTS}
As already discussed, the application of the aforementioned post-hoc XAI techniques
is limited when time-series data is involved. Especially, in MTS, finding meaningful constructs in high-dimensional information is not trivial. To deal with such data, segmentation techniques were mostly needed, splitting and analyzing data into valuable sub-sequences. 

More specifically, LIME and SHAP extensions were necessary to be further introduced, by leveraging the strong foundations of these methods adjusted to address such directions. For instance, TS-MULE is a LIME-based method not limited to uniform segmentation using static windows, but incorporating multiple advanced segmentation algorithms, such as matrix profile and SAX \cite{ts_mule}. This way, it is more likely that meaningful subsequences are uncovered, leading to motifs and local trends that are easily interpretable. Such methods can be also adapted to interpret forecasting output, instead of only dealing with the classical classification targets. 

In the case of SHAP, multiple extensions have been proposed and adjusted to a time-series setting. An example is an extension of KernelSHAP, adapted to explain time-series models, such as AR, ARIMA, VAR, VARMAX \cite{varmax}. This method focuses on computing feature importance values for time series data.
Another is the TimeSHAP method, aiming to explain more complex RNN-based models \cite{timeshap}. TimeSHAP provides explanations on multiple levels, by computing feature-, timestep-, and cell-level attributions.

\subsubsection{Adapting MTS as an input to classification models}
Another challenge of an MTS classification task arises from the complexity of time-series, which is not straightforwardly input to a classical machine learning model. The problem is the complex nature of time-series that makes it deviate from the conventional feature-vector representation. In the context of MTS, data is defined in a multi-dimensional feature space and characterized by special connections between the instances (time-points) as well as features. Thus, potential approaches focus on how to input the time-series data into the classical classification models.

The most straightforward way is to neglect any temporal dependencies by assuming instance and feature independence. This discards any time-oriented association and perceives the data directly as a vector input. Likewise, any classification model can be used on this dataset, but using the same output for all instances of the same time-series. Moreover, a well-applied approach addressing such data is to transform the complex-structured time-series data to a simple feature-vector representation. Such transformations can be achieved by using statistical-based or shapelet- and subsequence-based characteristics of the time-series \cite{time2feat}. After such transformation, a feature-vector representation is retained, which can be easily used in all existing classification models.

Alternatively, the necessary transformations can come internally through models incorporating data representations and prediction. Recently,  neural networks capable of handling multivariate time-series data have been used increasingly. More specifically, recurrent neural networks- (RNN) -based models, such as long-short term memory (LSTM) and gated recurrent unit (GRU) represent the state-of-the-art group of models in tasks involving sequential decision-making. Besides these, attention mechanisms have been also introduced to sequential modeling with the ability to uncover and highlight the most important parts or periods of sequences \cite{vaswani2017attention,hsu2019multivariate}. In other words, attention-based models offer interpretability for the results by using the learned attention weights. Utilizing attention weights is considered a form of inherent interpretability that is not commonly observed in NNs. 

According to the attributes of attention-based models, our two challenges, handling MTS data and MTS interpretability, seem to be overcome. Thus, our methodology is specifically designed to enhance the explainability of MTS clustering by employing an attention-based mechanism, strategically integrating both temporal and feature-level attention.

\section{Framework for Clustering Explanations}
In this section, first, the structure of the examined EMA data is described along with the problem statement of our work. Specifically, the goal of providing explanations on clustering results applied to EMA MTS data is presented in detail. Then, the focus will be on the proposed architecture that utilizes an interpretable attention-based framework and how this can eventually lead to clustering explanations.

\subsection{EMA Data}
In the current study, EMA was used as a dynamic approach to collect data on individuals' psychopathology-related behaviors, experiences, and symptoms in real-time and in their natural environments. More specifically, through digital questionnaires on their smartphones, participants were prompted to answer sets of questions. Most questions were answered on 7-point Likert scales; after normalization at an individual level, these data can be considered continuous. As all this data was collected multiple times throughout the day, EMA is typically structured as MTS data.
In this study, the dataset examined consists of real-world EMA pilot data, collected from students of Dutch Universities \cite{roefs2022new,martinez2023developing}. More specifically, 187 participants were asked to complete 8 questionnaires per day at fixed time intervals for 28 days. %As it is not feasible to include all variables separately in the analyses, 
According to the domain experts, some variables (e.g., positive affect and negative affect) were averaged across relevant variables (e.g., happy, calm, etc. and sad, angry, etc., respectively), and some variables were not included in the analyses, either due to limited within-person variance or due to relatively less relevance.

The EMA dataset, consisting of the data of $N=187$ individuals, can be denoted as $X=\{X_1, X_2,..,X_N\}$. For each individual $i$, $X_i$ is comprised of $V=12$ variables, each captured for a number of time-points, $T_i$. The total number of time-points $T_i$ ranged from 112 to 224 and was different across individuals due to differences in compliance. The minimum compliance was set to $50\%$. Each individual EMA dataset $X_i$ is represented by an MTS dataset, structured as
$\{V_1,\ldots,V , T_1,\ldots,T_i\}$.

\subsection{EMA Clustering}
As already stated, an important task in EMA analysis is clustering individuals based on their MTS data. Clustering entails assigning individuals sharing similar patterns to the same groups, while assigning individuals with different patterns to different groups. This approach enables uncovering distinct profiles among people that might not be immediately apparent. Such information could also provide insights into mechanisms maintaining or causing psychopathology.

A variety of clustering methods can be applied to EMA MTS data, each highlighting different underlying assumptions about the structure of the data~\cite{ntekouli2023evaluating}. These include traditional clustering techniques, such as k-means and hierarchical clustering, but are adapted to handle the temporal and multivariate nature of EMA data. For instance, clustering methods by incorporating an appropriate distance metric, like dynamic time warping (DTW) and Global Alignment Kernel (gak), are considered capable of advancing the inherent temporal dimensions of the EMA data ~\cite{cuturi2007kernel,cuturi2011fast}. Alternatively, clustering on EMA data could be approached from a multivariate functional data perspective, by representing the data with smoothed curves. The use of smoothing techniques could possibly facilitate identifying the underlying trends, which might be obscured by noise in the original time series \cite{jacques2014functional}. However, its applicability to EMA data should be explored more carefully. 

\subsection{Proposed Framework for Clustering Explanations}
\begin{figure}[t]
    \centering
    \includegraphics[width=0.9\textwidth]{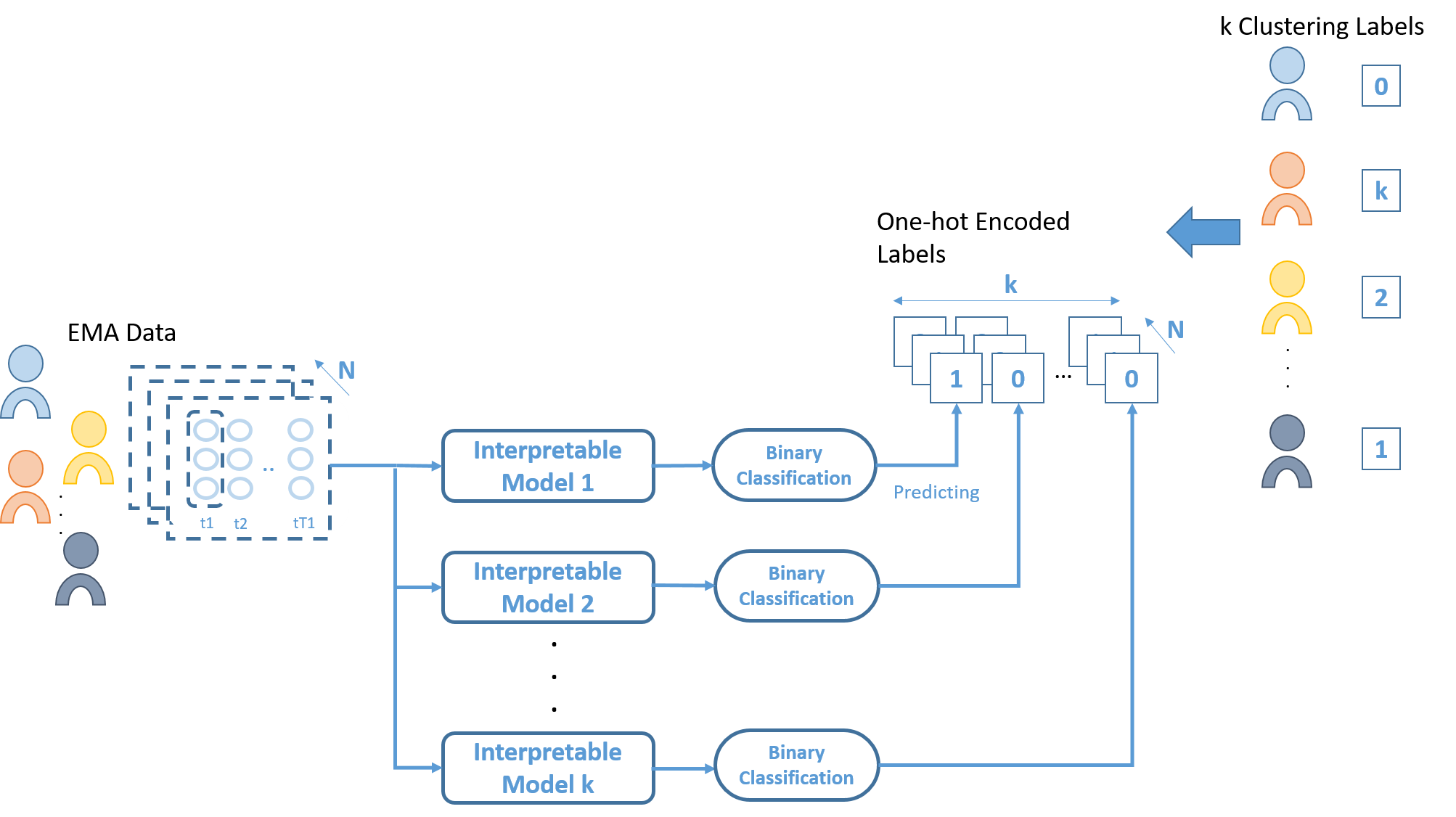}
    \caption{An overview of the proposed framework for providing explanations on EMA clustering.}
    \label{fig:framework}
\end{figure}

An overview of the proposed framework of providing explanations on EMA clustering is given in ~\ref{fig:framework}. All components of the framework are described as follows.

\subsubsection{Input: EMA Data}
The framework begins with the EMA data matrix $X$ as input. $X$ represents the personalized data of $N=187$ individuals, consisting of $V=12$ variables and $T_i$ time-points. Due to the variability observed in missing observations of individuals, we implement a padding strategy to complete the examined EMA dataset so the dimension of X is $\{187, 12, T\}$. Thus, each individual's data is transformed and filled when necessary, so that all have the same $T$ time-points for further analysis.

\subsubsection{Output: Clustering Labels}
In this study, we adopt an approach focusing on the explanation of clustering outcomes in EMA data without depending on a particular clustering method. That is, the approach relies on the fact that clustering is performed as a prior step, utilizing only the derived clustering labels in our framework to further provide explanations. An obvious advantage of this is the flexibility in the choice of clustering technique. Thus, as clustering is not an integrated component of the framework, it basically utilizes the clustering labels as the output to understand the reasoning behind the formation of each cluster.

\subsubsection{$k$ Interpretable Models}

Moving to the actual components of the proposed framework, in the case of $k$-clustering, it consists of $k$ interpretable classification models. Each model focuses on predicting one cluster and distinguishing this from the rest of the clusters. For example, regarding the classification Model0, where the goal is to predict Cluster0, the output labels are 1 for individuals of Cluster0 and 0 for Cluster1 and Cluster2. By predicting a single cluster through an interpretable model, we could have access to the description of the model highlighting the special characteristics, dynamics, and features that differentiate that cluster from the rest. Then, fitting $k$ models could facilitate explaining all $k$ derived clusters, which means exploring the important variables and time-points for each cluster.

Practically, although the input of each model is the whole EMA dataset, the difference of the $k$ models comes into the output. This works by one-hot encoding the clustering labels of the pre-applied clustering and using a different output vector for each model. This encoding method transforms the categorical cluster labels into a binary matrix, ultimately forming $k$ binary classification models. It should be noted that in the case of a 2-clustering, the framework consists then of only one interpretable model. As the clustering output is already binary, there is no point in fitting two models.

\subsubsection{Cluster-specific Binary Classification Model}
\begin{figure}[t]
    \centering
    \includegraphics[width = \textwidth]{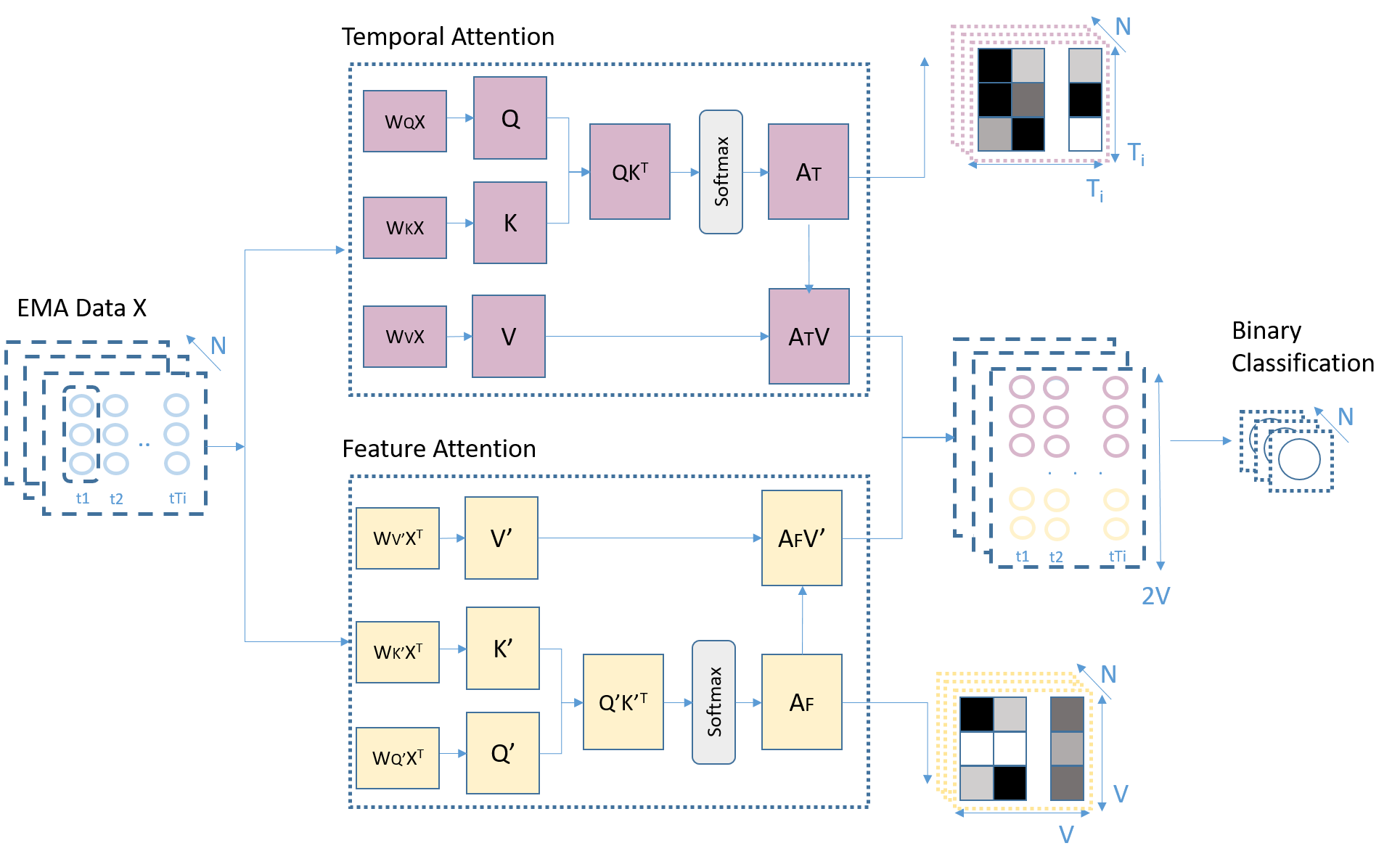}
    \caption{The main components of each interpretable model.}
    \label{fig:model}
\end{figure}

As already discussed, each binary classification model aims at predicting the individuals of a single cluster over the rest of the clusters. The main components of each model used are shown in Fig.~\ref{fig:model}. Each binary classification model is interpretable mainly relying on self-attention mechanisms \cite{vaswani2017attention}. By nature, the attention mechanism focuses on the temporal domain of the data, identifying the most important parts of the data with respect to the prediction task. By keeping only the relevant parts of the data and minimizing or filtering out the effect of irrelevant ones, the input space is effectively sparsified.
This is implemented by setting different attention weights on each time-points depending on the contribution to the output. Therefore, such weights could be beneficial in recognizing the important parts of the data for classifying an individual in one cluster \cite{hsu2019multivariate}.

In the current framework, to address the complexity of EMA data, 2-level attention is used in parallel, each focusing on different aspects of the data\cite{vskrlj2020feature,hsieh2021explainable}. The first level of attention is dedicated to uncover the important parts in the temporal dimension, while the second one is to the important features. The value of analyzing data at the feature level is particularly evident when it comes to interpretation, as it is inherently more insightful to offer explanations based on specific features.

The description of each attention-based mechanism is as follows:

\begin{itemize}
 
\item{\textbf{Temporal Attention}}

As shown in Fig.~\ref{fig:model}, the calculations of the Temporal Attention follow the procedure of a Scaled Dot-Product Attention self-attention block \cite{vaswani2017attention}. First, the 3 main components of the attention block, matrices query $Q$, key $K$, and value $V$, are calculated by a linear transformation of the input $X$ data. Then, the product of $Q$ and $K$ is the attention matrix, which after a Softmax normalization produces the actual matrix of temporal attention $A_T$. For each individual $i$, the dimensions of $A_T$ is $T_i \times T_i$. Subsequently, the attention matrix is multiplied by the matrix $V$ (linearly transformed input $X$) to yield the context vector. In other words, the context vector is the weighted input based on the learned attention weights.

By learning the weights in $A_T$, the temporal attention component is designed to identify the most significant time points within the EMA data. These weights are different for each individual and each time-point resulting to a $T_i \times T_i$ matrix. This matrix captures the relative importance of each time point with respect to every other time point, offering rich insights into temporal dynamics. However, its size and complexity pose challenges for direct interpretation and visualization. Despite the detailed and informative structure of the derived $A_T$, we employed a strategy of averaging over one dimension, while ensuring the other dimension is normalized. This approach averages the contributions of all time-points on each one, yielding only the average effect over time. This significantly simplifies the attention matrix, retaining only one dimension of size $T_i$ or a time-series of $T_i$ time-points.

\begin{figure}[t]
    \centering
    \includegraphics[width=0.7\textwidth]{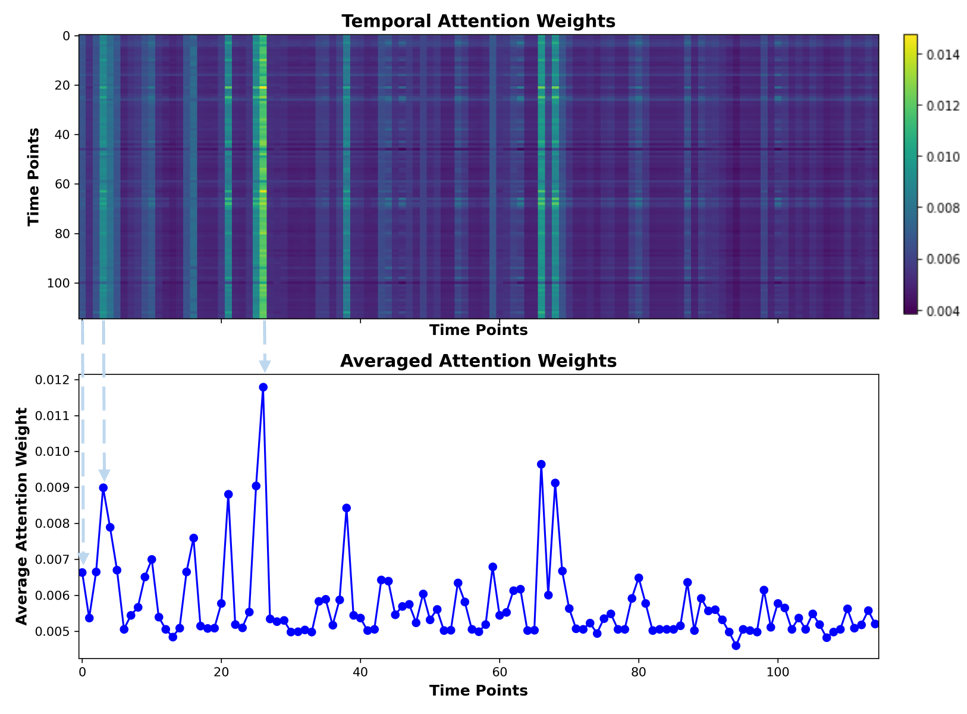}
    \caption{The averaging process of the full Temporal Attention matrix $A_T$ to $A_{T_{av}}$}
    \label{fig:average}
\end{figure}

\item{\textbf{Feature-level Attention}}

In parallel to temporal attention, the feature-level attention mechanism assesses the importance of each feature within the EMA data. As noticed in Fig.~\ref{fig:model}, the block of feature-level attention is the same as of the temporal attention. The only difference is that the initial $Q’, K’$ and $V’$ matrices come after a linear transformation of the transpose of the input EMA data, $X^T$. As before, this component results in the attention weight matrix $A_F$ and the context vector. While the latter has the same size as the input $X$, the feature-level attention is in the dimension of $V \times V$. This is designed to map the inter-feature relationships and their contributions to the model's predictions. Through such weighting, the framework can distinguish which variables play an important role in influencing the outcome, providing an additional layer of explanation that complements the temporal insights. 

Regarding interpretability, it is evident that being in the feature domain makes it more straightforward to provide a deeper understanding of the role of influence for each variable. Additionally, the low feature dimensionality allows us to directly represent, visualize, and analyze the feature-level attention weights. This offers clear insights into which features are the most relevant and how these features influence each other within the context of the prediction task. 
\end{itemize}

\subsubsection{Clustering Explanations through Attention Weights' Analysis}

To provide clustering explanations, the produced averaged temporal attention weights and feature-level attention weights are further analyzed. The weights’ analysis first facilitates the description of each model, aiming to differentiate each cluster, and then explaining the observed differences. In the following analysis, we adopt a multi-aspect approach to present our findings, covering a thorough exploration of the impact that attention mechanisms have at various but interconnected levels. These involve the cluster- and individual-level. From high- to low-level, each level zooms into different parts of the data, offering unique insights into the underlying relationships.

\begin{itemize}
    \item \textbf{Cluster-level Analysis}
    
At the cluster level, the focus is on separately describing the individuals belonging to each cluster to get a clearer picture of the group-specific behaviors and attributes that distinguish one cluster from another. By examining the averages of attention weights across all individuals of each cluster, derived from the model describing each cluster, the aggregated behaviors and patterns of a cluster can be identified. At a high level, the average weights could uncover the special characteristics or strong effects that all these people have in common and possibly drove them to belong to the same cluster. 

Although both temporal and feature-level attention weights are explored, getting access to the average effects across individuals based on the full temporal attention $A_T$, or even the averaged temporal attention $A_{T_{av}}$, is a bit challenging. As already discussed and shown in Fig.~\ref{fig:average}, each individual is described by a time-series, $A_{T_{av}}$, showing their important time-points. However, it is not meaningful to average over different time-series, because the important time-points differ across individuals. To address this, the correlation between the attention weights $A_{T_{av}}$ and each feature's time-series is calculated, potentially identifying which features consistently align with the attention temporal trend. 
Eventually, at this level of analysis, through both the temporal and feature-level attention, the importance of specific variables in leading to a particular cluster-output could be uncovered.

Subsequently, at a cluster-level, the patterns of inter-variable relationships or interactions dominant for each cluster are investigated. Based on the model’s description through the attention weights, the relationship between data across individuals of one cluster can be also checked with respect to the acquired weights.

\item \textbf{Individual-level Analysis}

At the individual level, the attention weights are examined separately for each individual, allowing for a personalized interpretation of the data. Without transforming or averaging the attention weights, the actual learned weights are analyzed along with the original feature space of each individual. Therefore, the focus is on understanding why or what was important to drive each individual to belong to a particular cluster. More specifically, regarding the time domain, the temporal attention weights facilitate uncovering what is happening underlying the time-points that are important for the prediction output. For instance, it is interesting to show which combinations of feature values get higher attention and which get lower.
\end{itemize}

\section{Analysis and Results}
As already mentioned, in our analysis, a real-world EMA dataset is used consisting of 187 individuals, 153 padded training time-points ($70\%$ of the total 224 time-points), and 12 distinct variables. After comparing different clustering methods based on different intrinsic evaluation measures, a 3-clustering result derived from a gak kernel k-means was chosen as a good clustering~\cite{ntekouli2023evaluating}. Then, the clustering labels derived from that particular method are further investigated to provide explanations regarding the formation of the clusters. Thus, the goal is to uncover what is different among clusters, meaning the important characteristics that drive each individual to belong to that specific cluster.

Taking as inputs the dataset $X$ and the cluster labels, the proposed framework for describing and explaining clustering can be employed. According to the structure of the framework, for a 3-cluster grouping, a set of 3 interpretable models is trained on all individual EMA data, each aiming to predict one cluster over the rest. Therefore, the labels are one-hot encoded and each one-hot vector is the output of one model. 

\subsection{Performance Evaluation}
Since the following analysis relies on the parameters derived from the 3 models, it is important to evaluate their effectiveness. The performance is then compared against the individual component of temporal attention and the baseline LSTM model. The assessment focuses on the ability of each approach to predict the cluster labels (provided by the pre-defined clustering) based on the EMA MTS data. After splitting the data into training (first 70\% of individual time-points) and test (last 30\%) sets, the number of correctly classified individuals (out of 187) is calculated. Each model was trained on the same training and test datasets, ensuring consistency in evaluation. The averaged results over the 3 models of each approach are presented in Table \ref{tab:perf}.

\begin{table}[t]
\caption{Comparison of Models Performance}
    \centering
    \begin{tabular}{lcc}
\hline
Model & Training Accuracy & Test Accuracy\\
\hline
     
Baseline LSTM                   &   144/187     &   100/187      \\
Temporal-Attention     &  159/187   &        100/187           \\
Proposed Framework   &  187/187                &   105/187       \\ \hline
\end{tabular}
    \label{tab:perf}
\end{table}

This comparison shows the superior performance of the proposed framework over both the baseline LSTM model and the individual attention-based component. By effectively integrating the temporal and feature-level attention mechanisms, all individuals are identified in the correct cluster in the training set. Therefore, it is expected that the weights of a better-performing model could more accurately reflect the description of the underlying prediction task, which is the clustering. The rest of the analysis is conducted on the training set of all individual data. 

\subsection{Cluster-level Explanations through Temporal-Attention}

To describe the characteristics of a cluster, the focus is on the model predicting that cluster label over the rest of the clusters. For example, for the first cluster, Cluster0, the analysis is conducted on the parameters of the first model, Model0. Regarding that model, the individuals of the two classes, which means belonging to one cluster over the rest of the clusters, are separately analyzed. Thus, by showing the average effects of the parameters of the individuals belonging to Cluster0, we could provide some description of Cluster0. A similar procedure holds when describing all different clusters. In other words, the parameters of Model1 are used for describing Cluster1 and the parameters of Model2 for Cluster2. Then, across all 3 models, the individuals belonging to the associated cluster, are separately analyzed. 

\begin{figure}[t]
    \centering
    \includegraphics[width=\textwidth]{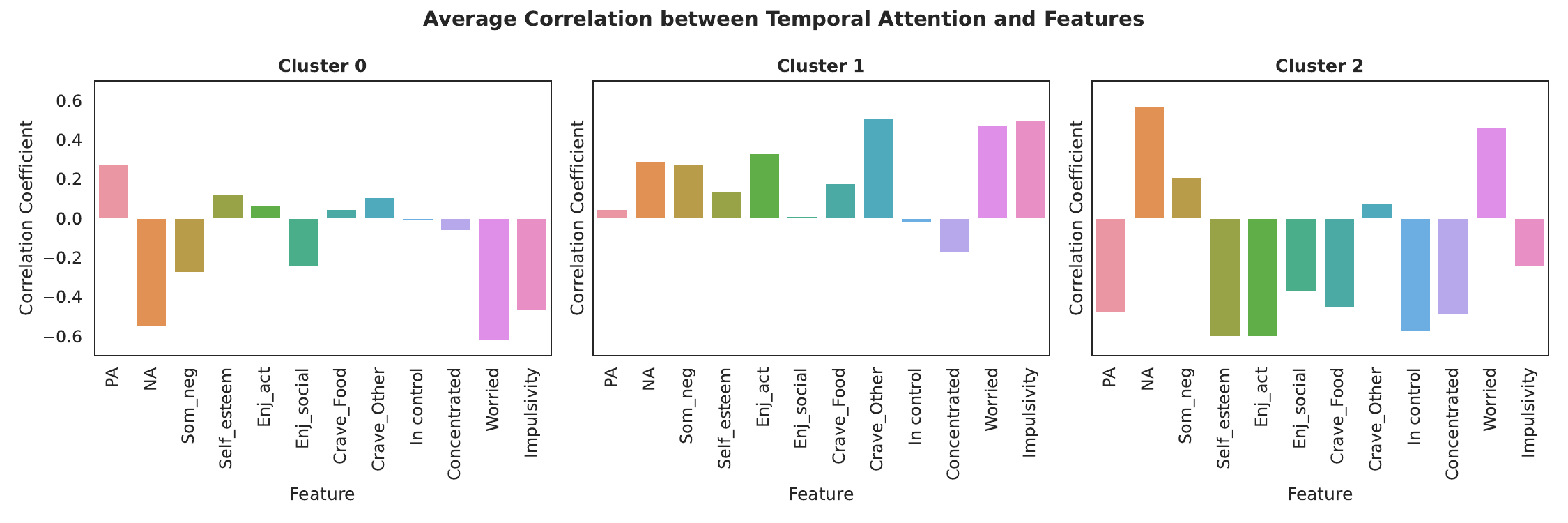}
    \caption{Cluster-level average correlation effects between the temporal attention weights and the EMA features.}
    \label{fig:TempCorr}
\end{figure}

As already discussed, to give an overview of the temporal effects on cluster-level, a correlation analysis is employed between the temporal attention weights $A_{T_{av}}$ and the time-series of each feature. The average correlation scores across all individuals of each cluster are presented in Fig.~\ref{fig:TempCorr}.

According to Fig.~\ref{fig:TempCorr}, the important effects of features in distinguishing each cluster from the rest are identified. It can easily be seen that the patterns of the correlations of EMA features and temporal attention weights are almost orthogonal for each of the clusters. Furthermore, particular high (low) correlations stand out. In Cluster0, ``Negative Effect" (given as NA), ``Worried" and ``Impulsivity" have strong negative correlations with, meaning that the high values of these features get a lower attention weight. In Cluster1, almost all derived correlations are positive, with the highest being for ``Craving\_Other", ``Worried" and ``Impulsivity", whereas in Cluster2, almost all correlations are stronger apart from the feature Craving\_Other. 
Although a cluster description has been uncovered by the analysis above, it still relies on the average (across all individuals of a cluster) effects. In other words, it is possible that some individuals' effects deviate from the averaged ones. Thus, explanations at the individual-level should be also further explored. 

\begin{figure}[t]
\centering
\begin{subfigure}[b]{\textwidth}
\includegraphics[width=1\linewidth]{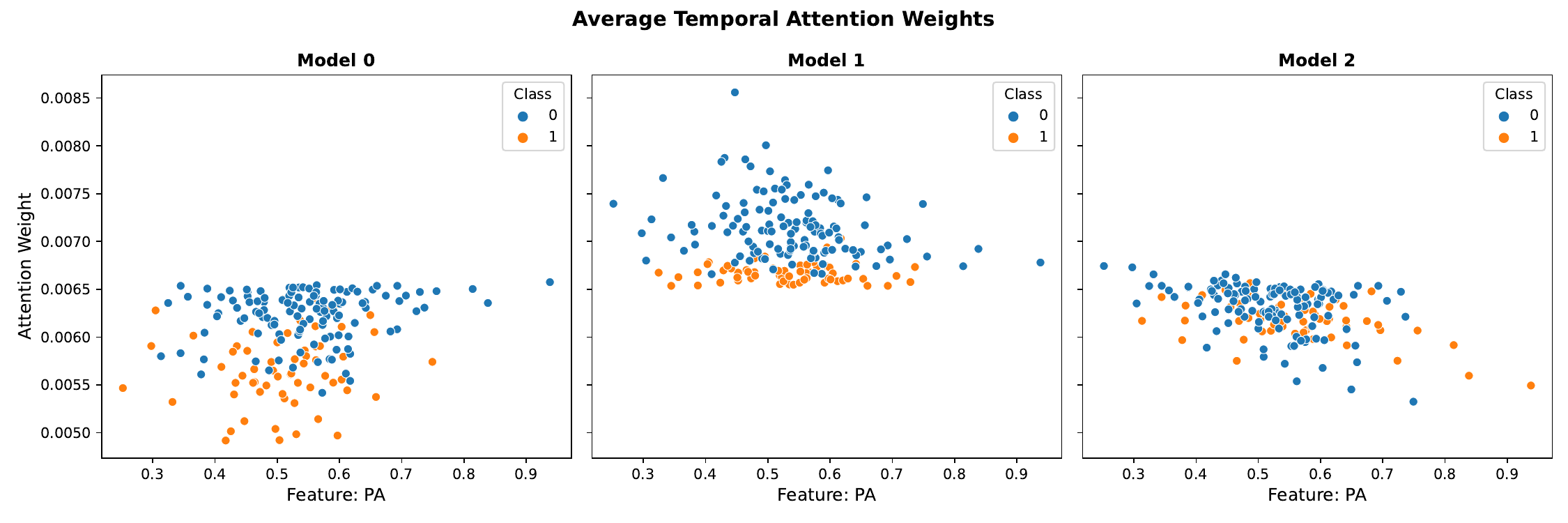}
\caption{Differentiation of attention weights across Class0 and Class1 of all 3 models.} 
\end{subfigure}

\begin{subfigure}[b]{\textwidth}
   \includegraphics[width=1\linewidth]{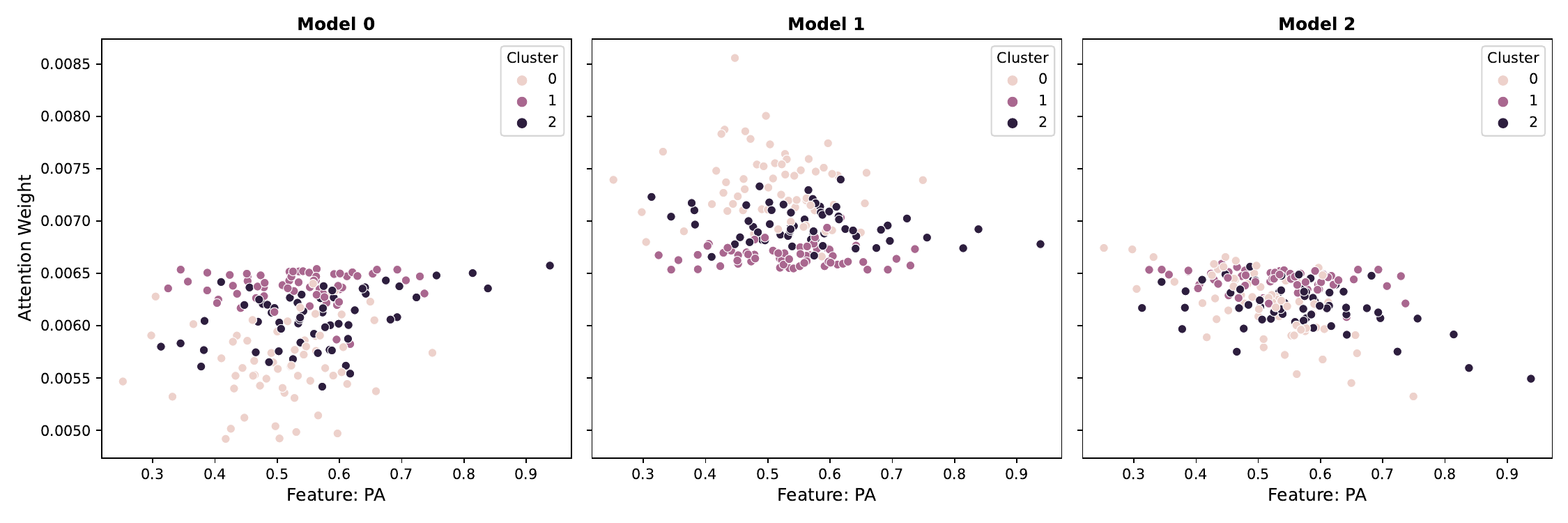}
   \caption{Differentiation of attention weights across the actual 3 clusters.} 
\end{subfigure}
\caption{Feature (Positive Affect) Relationship with Temporal Attention weights}
    \label{fig:Labels}
\end{figure}

Despite the first findings on clusters' composition, the actual role of attention has still been unclear. For instance, we need to understand what it actually means to get a higher or lower attention score. Therefore, we could analyze the average attention weights of individuals belonging to each cluster (represented by Class1 in each model) or those not belonging to that cluster (represented by Class0). As before, across all models, the average attention weight of all 187 individuals is shown in Fig.~\ref{fig:Labels} with respect to their average values of one feature. In each sub-figure, all 187 individuals are shown, represented by a point and colored according to their output class in each model. Regarding the coloring, these are different on the first and second row of the figure: while on the first row, the real class label (0 or 1) is depicted, on the second row the actual cluster label. 

We notice that for Model0 and Model1, individuals belonging to Cluster0 and Cluster1, respectively, get lower (on average) attention scores than the rest. This could possibly reflect the fact that Class1 is always the minority class compared to Class0. Thus, each model gives more attention to the majority class. Nevertheless, the findings of Model3 are not similarly clear. It is noticeable that the attention weights of the Cluster0 and Cluster2 are slightly mixed, but getting lower values than Cluster1. These unclear results should be further investigated as it may show that the third cluster may not be needed.   

Additionally, no significant effect is observed on the feature-level. We can see that individuals with average values ranging from 0.3 to 0.9 can belong to all possible clusters. This is also apparent when plotting across any other feature, where the same patterns are found. 

\begin{figure}[t]
    \centering
    \includegraphics[width = 0.49\textwidth]{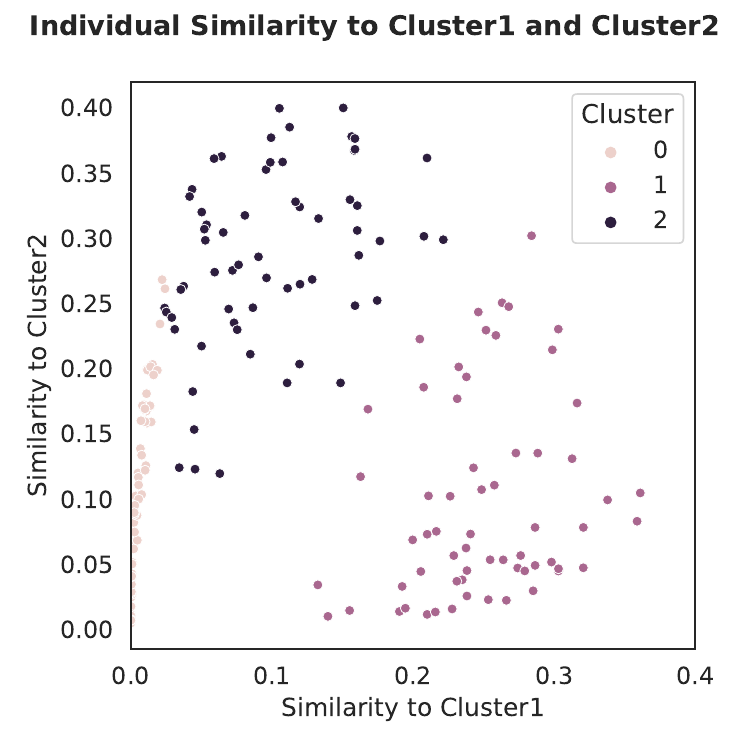}
    \caption{Similarities of all individuals to Cluster1 and Cluster2.}
    \label{fig:dist}
\end{figure}

In response to the overlapping attention weights of Cluster0 and Cluster2, the similarities across clusters should be analyzed. Since our clustering relies on the gak similarities, these are plotted for all individuals. In Fig.~\ref{fig:dist}, the similarities of all individuals (colored by their true cluster) are depicted in Cluster1 and Cluster2. Practically, for each individual, the average similarity to all the individuals of each cluster is given. Because of a 2d plot, the similarities to Cluster0 are not shown. 

As expected, it is interesting to see that there is a similarity between Cluster0 and Cluster2. Although individuals of Cluster0 have very low similarity to Cluster2, lower than 0.05, the majority of them have a similarity between 0.10 and 0.30 to Cluster2. This range is not far from the within-cluster similarity which only reaches the level of 0.40. In addition, we can observe individuals of Cluster1 and Cluster2 exhibiting comparable levels of similarity to more than one cluster, suggesting a degree of ambiguity in their cluster membership. Such findings raise questions regarding the quality and robustness of the chosen clustering. Therefore, it is indicated that the current clustering method may not fully capture the underlying heterogeneity of the dataset.

\begin{figure}[t]
    \centering
    \includegraphics[width=\textwidth]{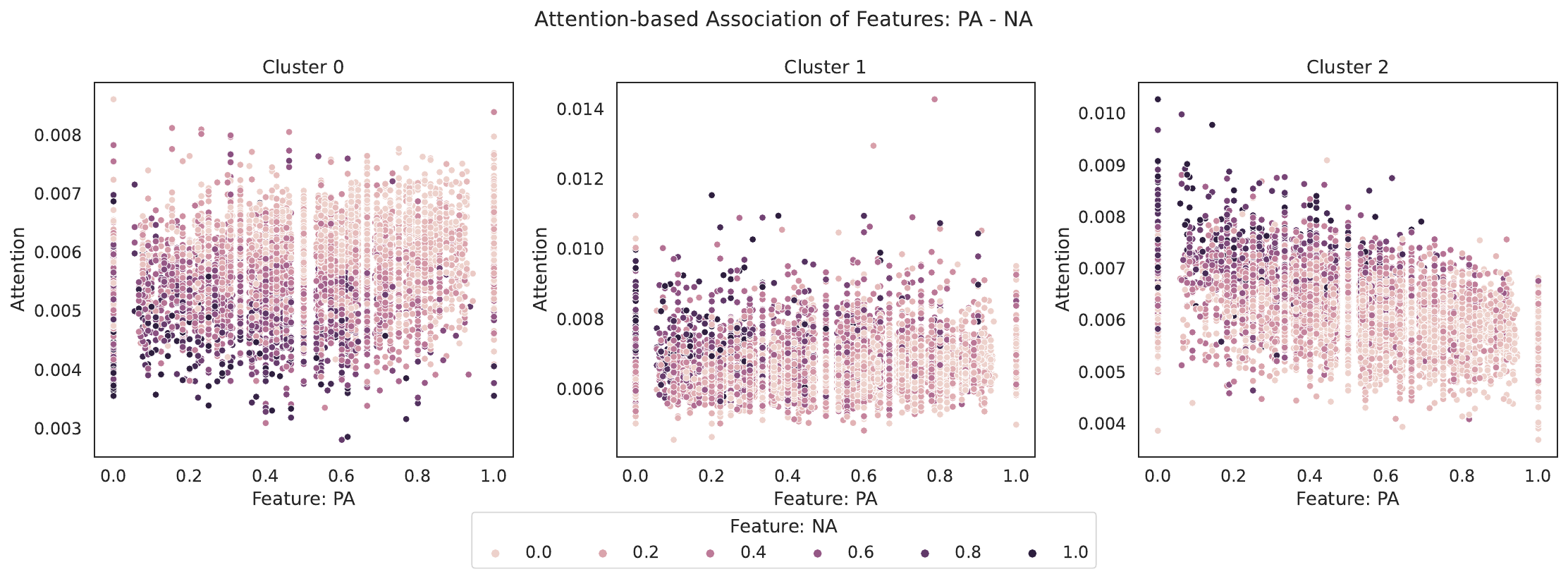}
    \caption{Features Interaction between ``Positive Affect" (PA) and ``Negative Affect" (NA) with respect to the Temporal Attention.}
    \label{fig:pa-na}
\end{figure}

In the context of cluster-level analysis, beyond the initial feature-related effects, we could further elaborate on the analysis using feature interactions. For example, the interaction between two features, ``Positive Affect" (PA) and ``Negative Affect" (NA), along with the acquired attention weight is shown in Fig.~\ref{fig:pa-na}. For each cluster, each point represents a time-point of the individuals belonging to that cluster. According to this figure, the different interaction patterns underlying each cluster can be identified. More specifically, for Cluster0, high NA values lead to low attention weights, whereas the opposite effect is seen for Cluster1 and Cluster2. Although for Cluster0 and Cluster1, no visible interactions were observed, a quite clear pattern is seen for Cluster2. The combination of low PA and high NA leads to high attention weights, whereas high PA and low NA lead to lower attention weights. By exploring all possible feature interactions, we have the opportunity to get more insights into the underlying structure of each cluster. Thus, the potential distinctions could point to differences in which variables are most relevant characterizing each cluster.

\subsection{Cluster-level Explanations through Feature-Attention}
\begin{figure}[t]
    \centering
    \includegraphics[width=\textwidth]{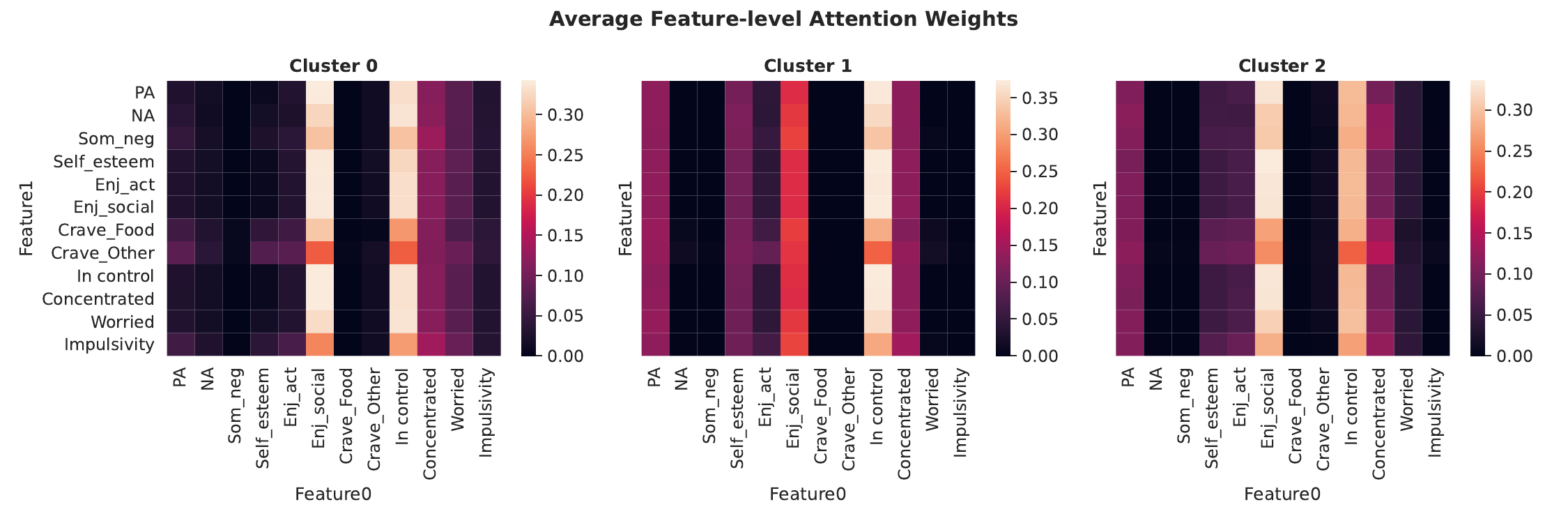}
    \caption{Cluster-level average feature-level attention weights.}
    \label{fig:FeatAtt}
\end{figure}

Regarding the learned feature-level attention weights, the raw $A_F$ weights are directly analyzed. Similarly to the previous section, to get some insights at a cluster-level, the average effects of all individuals within each cluster are aggregated. The average feature-level attention weights for each cluster are given in Fig.~\ref{fig:FeatAtt}. The $12 \times 12$ heatmap presents the asymmetric inter-relations among features, specifically showing the relative contribution of Feature0 (x-axis) to Feature1 (y-axis).   

\begin{figure}[t]
\centering
\begin{subfigure}[t]{\textwidth}
   \includegraphics[width=1\linewidth]{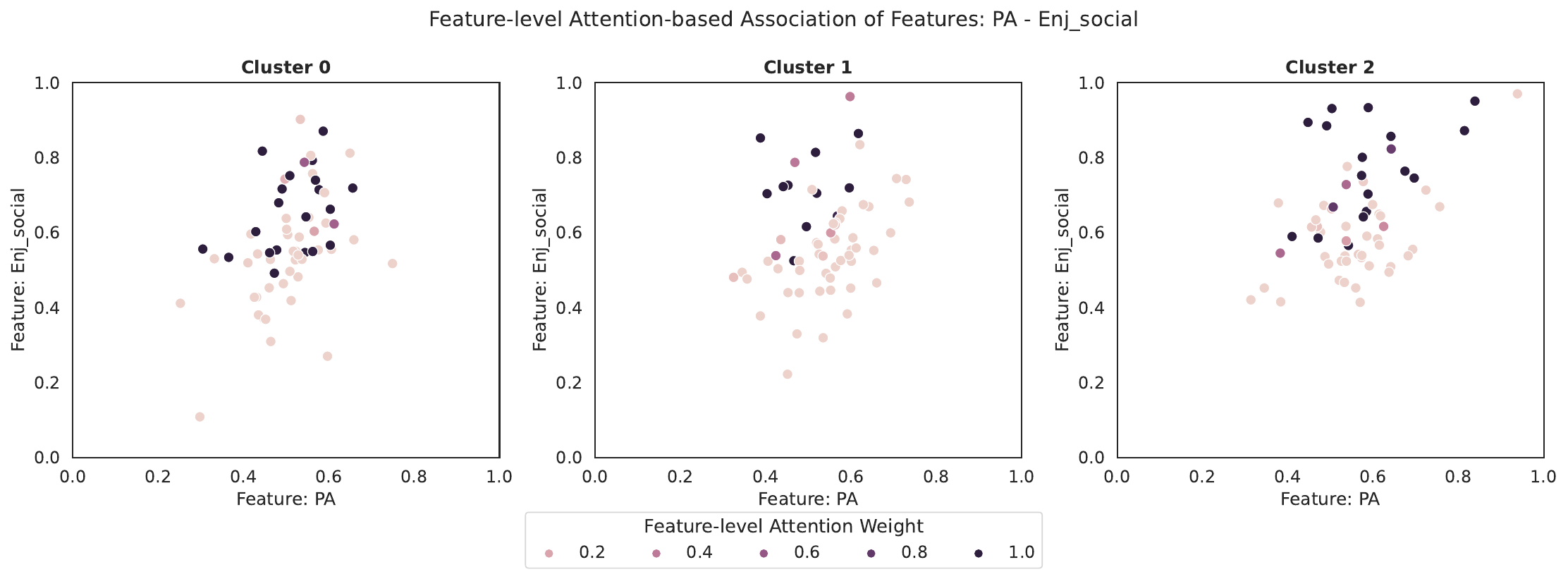}
\end{subfigure}

\begin{subfigure}[t]{\textwidth}
   \includegraphics[width=1\linewidth]{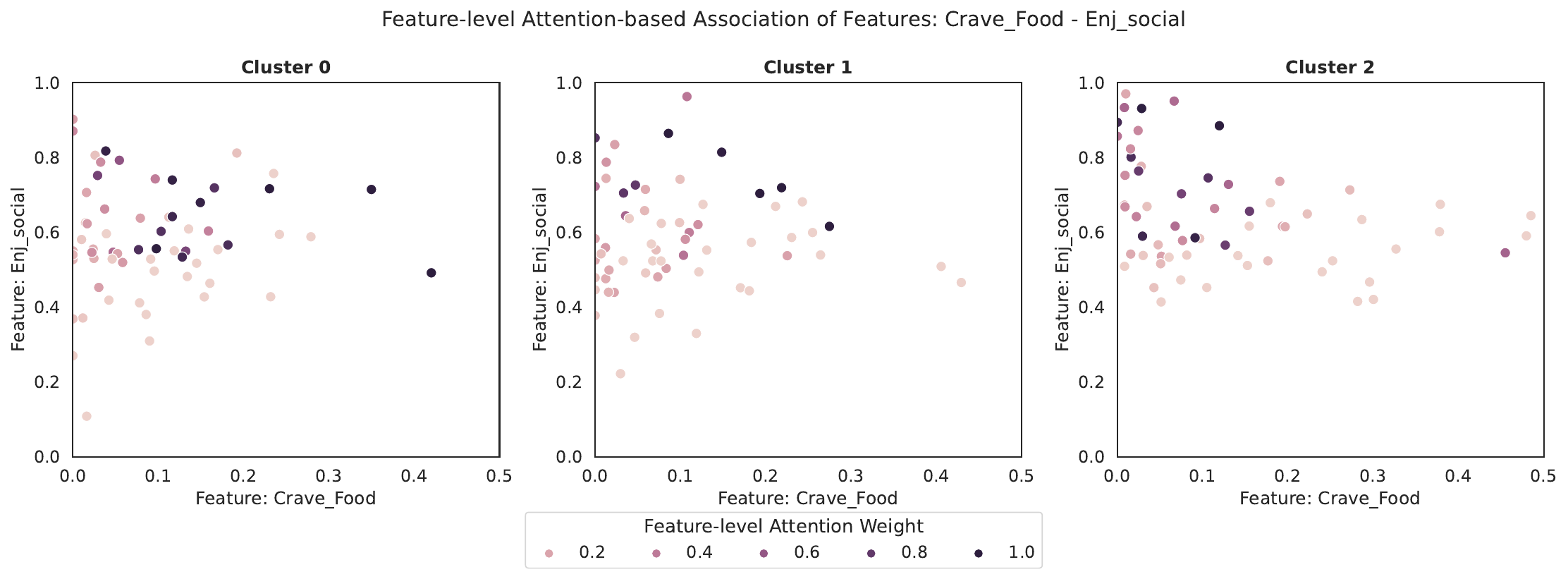}
\end{subfigure}
\caption{Feature-level Attention: Unfolding the inter-connection of ``Enjoying Social Activities" to two other features: ``Positive Affect" and ``Crave Food".}
    \label{fig:feat5}
\end{figure}

The averaged attention weights can be similarly interpreted as the relative importance of various features in distinguishing one cluster from the rest. At first glance, it's noticeable that the same features emerge across all clusters, ``Enjoying Social Activities" and ``In Control", but to a different importance degree. For instance, this means that high values of ``Enjoying Social Activities" are assigned to high attention weights in relation to any other variables. To illustrate this, an example of these interconnections is given in Fig.~\ref{fig:feat5}. More specifically, the association of the average ``Enjoying Social Activities" value with the average ``Positive Affect" and ``Crave Food" across all individuals within each cluster is depicted in the first and second row, respectively. Each point represents an individual belonging to a particular cluster. As expected from Fig.~\ref{fig:FeatAtt}, we can identify the high attention contribution of high values of ``Enjoying Social Activities" to the other features along with other interesting patterns. For instance, in Cluster2, high attention weights are assigned for low values of ``Crave Food" and high values of ``Enjoying Social Activities".
By analyzing such feature interplay across all combinations, from the perspective of feature-level attention, deeper insights into the underlying dynamics can be uncovered. Thus, a more detailed examination of feature interactions could reveal some special characteristics and patterns that are important for distinguishing each cluster.

\subsection{Individual-level Explanations}
While cluster-level analysis offers valuable insights into the commonalities within clusters, undoubtedly it smooths over individual differences. To more thoroughly understand the important patterns of cluster formation, it is essential to examine the data at the individual level as well. 

\begin{figure}[t]
    \centering
    \includegraphics[width=0.8\textwidth]{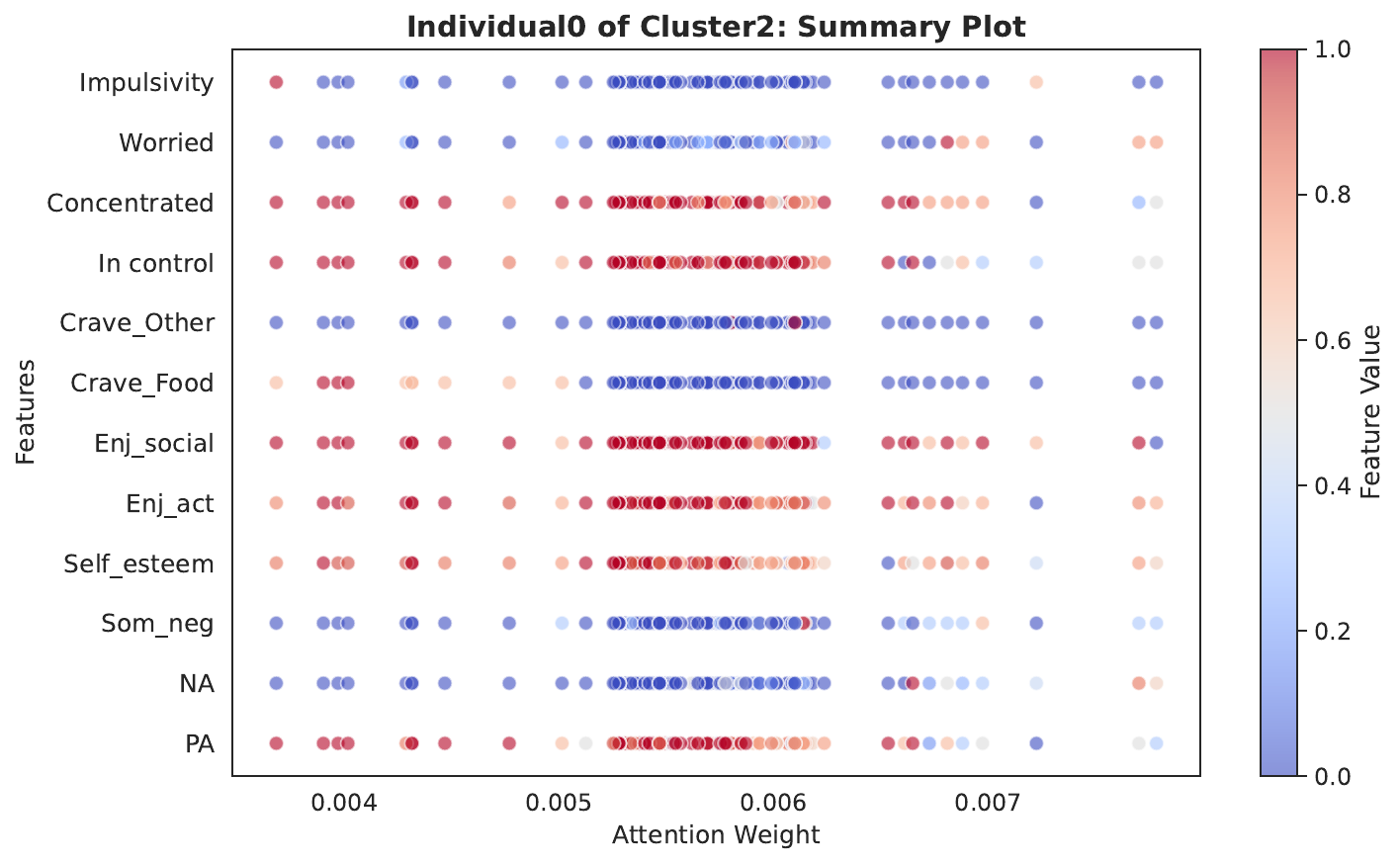}
    \caption{The summary plot of feature interactions for an individual of Cluster2.}
    \label{fig:ind0_sum}
\end{figure}

Beyond unfolding the individual effects of the cluster-level analysis conducted before, it is important to investigate the underlying interactions of features in response to the attention weights learned by the model dedicated to the cluster each individual belongs to. The example of the first individual, Individual0, belonging to Cluster2, is used for the rest of the analysis. A detailed summary plot including all feature interactions and their attention weights is presented in Fig.~\ref{fig:ind0_sum}. Particularly, for each feature on the y-axis, all the time-points of Individual0 are plotted, while ordered according to their attention weights, and colored by the value of each feature. Through this, we can have a thorough exploration of the combinations of feature values that lead to higher attention. According to the model, higher attention weights show the important characteristics for distinguishing individuals between clusters. Although it was indicated before that on average low attention was assigned to individuals classified as Class1, which corresponds to Cluster0, this pattern was not that dominant for Cluster2. Thus, further investigation of all individuals is necessary for a deeper understanding of the distinct feature dynamics, especially in the case of Cluster2.

\begin{figure}[h]
    \centering
    \includegraphics[width=\textwidth]{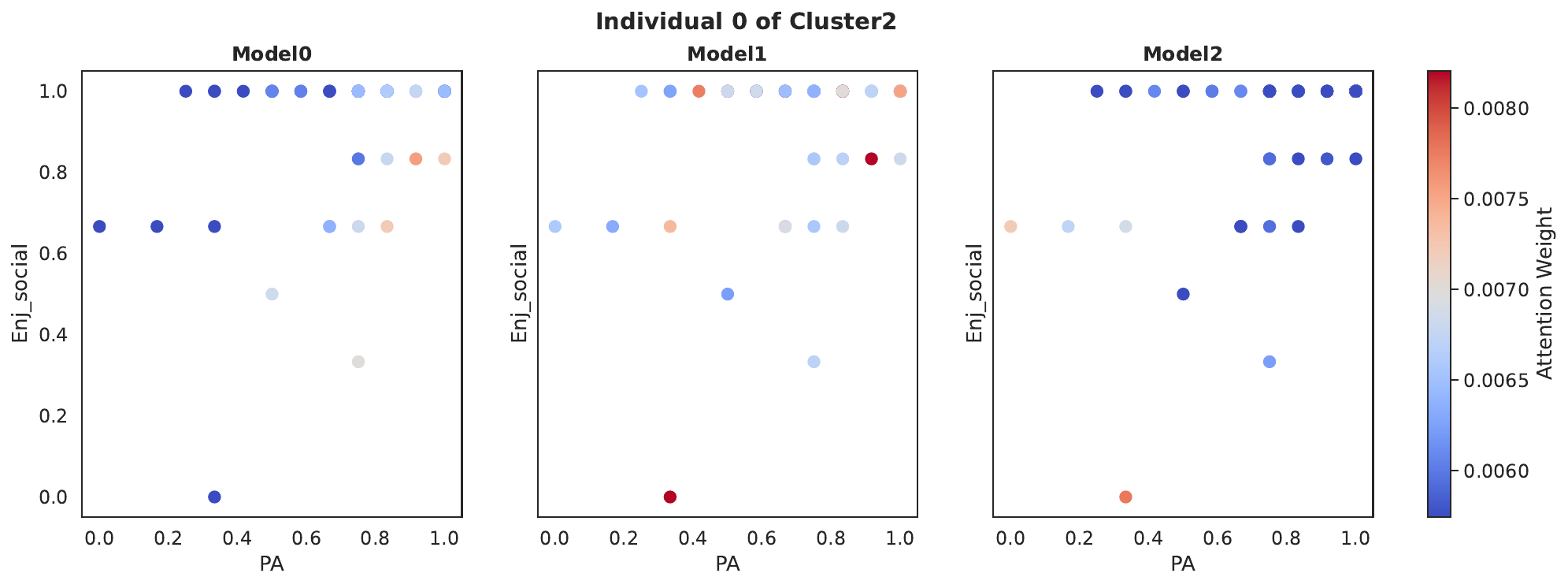}
    \caption{The attention weights of the interaction between ``Positive Affect" and ``Enjoying Social" derived from all 3 models.}
    \label{fig:ind0_feat}
\end{figure}

To more clearly understand the role of attention on each individual, the differences in the weights learned by all models should be additionally studied. Specifically, for each individual, feature interactions could be compared against the attention weights derived from each model. This could facilitate uncovering how much each feature interplay influences each model's decisions for each individual. An example of an individual belonging to Cluster2 is shown in Fig.~\ref{fig:ind0_feat}. By examining how the same feature interactions are colored based on the 3 models, some distinctions are apparent. For example, it is observed that the combination of high values in both ``Positive Affect" and ``Enjoying Social" leads to higher weights for Model0, whereas to lower ones for Model2. Thus, these patterns possibly reflect the impact that the specific time-points have on predicting Cluster0 and Cluster2, respectively. Also, after the previous indication that lower attention weights are linked to a particular Cluster, the lower attention of Model2 could highlight the most relevant information for predicting that individual as Cluster2. Therefore, such a comparison offers valuable insights into how different models prioritize and interpret the same set of features of an individual. 

\section{Discussion}
In this paper, the under-studied problem of providing explanations on clustering results in the context of MTS data is explored. A novel interpretable framework is proposed and examined using a real-world EMA dataset. To address the complexity of EMA data, our framework offers interpretability by integrating 2 levels of attention mechanisms, both in the temporal and feature-level dimension. Through its multi-aspect design and analysis, this framework eventually facilitates a deeper understanding of clustering, providing interpretations of the important underlying patterns. Next, we focus on the role of the multi-aspect framework’s design and analysis and the impact of prior applied clustering on interpretation and its validation.

\subsection{The Role of the Multi-aspect Attention}
Although attention mechanisms have typically been employed on the time dimension, the current approach of integrating multi-aspect attention mechanisms, focusing on both temporal and feature levels, showed that enhances the performance of the overall model in terms of accuracy. Based on Table \ref{tab:perf}, this integrated approach outperformed all baseline methodologies, by predicting more individuals in the correct clusters in both training and test sets. Its high performance is indicative of how informative the weights learned by each model are and how these could facilitate distinguishing individuals across clusters.

Moving to interpretability, the multi-aspect attention mechanism offers detailed information on the data patterns that each model considers as important. While temporal attention highlights the significance of particular time-points, such information remains meaningful on the individual level since each individual exhibits behaviors at different time-points. To enrich the learned information and our understanding on cluster-level, feature-level attention can additionally identify the important interconnection among features. Thus, incorporating both temporal and feature-level attention enhances the interpretability of complex data like EMA, where the understanding of dynamic patterns is crucial. 

\subsection{The Role of the Multi-level Analysis}
The analysis of all learned attention weights is conducted at 2 levels, cluster- and individual-level. The cluster-specific insights provide a deep investigation of the commonalities as well as distinctions across clusters. After aggregating the effects of all individuals in each cluster, we derived the most influential time-points as well as the underlying features and feature interactions based on a model’s decision-making. Although some first cluster descriptions were derived, it should be noted that the average effects smooth over the real individual differences. Therefore, to better understand each cluster formation, it is essential to examine the data at the individual level. The individual-level analysis facilitates uncovering important feature interactions and patterns unique to each subject as well as how each person is reflected on the weights learned by models focusing on different clusters.

\subsection{The Impact of Utilizing a Pre-defined Clustering Result}
A key advantage of the framework is its algorithm-agnostic nature regarding the generation of clustering labels. By design, as clustering is not integrated into the framework, it shows great flexibility allowing every possible clustering technique to be potentially prior utilized. Thus, it could be used as an effective tool for evaluating the result given by any clustering method. Such a feature not only enhances the framework's utility but also broadens its applicability, allowing for the comparison and evaluation of clustering algorithms based on the quality and relevance of the insights they produce.

However, it is reasonable that the quality of the clustering results utilized in the framework plays an important role in the accuracy of the explanations. In other words, in the case that clustering labels are not robust, the proposed framework would again provide some explanations on the cluster- and individual-level importance effects, but without these being meaningful or accurate.
This may be the case also for the chosen clustering of our EMA dataset. Since the real clusters are not available and the between-cluster similarities are found close to each other, the examined clustering may not be the optimal one. This can be caused by the clustering algorithm utilized or the complexity of the data. Nevertheless, even when using that clustering, the framework was capable of uncovering similarities in the influential data patterns of different clusters, suggesting that a 2-clustering result may be more possible. Therefore, regarding the evaluation of this framework, it is important for the clustering results of other clustering algorithms to be investigated. This necessity also comes since in human-centric fields, like EMA studies, developing simulation datasets or objective criteria to evaluate the quality of explanations is quite challenging. Thus, explanations should further be assessed by domain experts on a case-by-case basis. Alternatively, explanations could be investigated using some different types of ground-truth information. This could include baseline data (such as demographic details or health-related information) collected before or during the study.

\section{Conclusion}
To summarize, our interpretable framework represents a significant advancement in providing explanations and evaluating an MTS clustering result. By thoroughly analyzing the attention-derived important time-points and feature interactions at both cluster- and individual-level, valuable insights into the patterns characterizing each cluster are identified. This is especially critical in fields like psychopathology, where a better understanding on clustered individuals could be beneficial for personalized interventions and mechanistic understanding.

Regarding future work, the goal is to augment personalized models by using data from similar individuals. Having evaluated various clustering results and uncovered the important data segments of an individual through the proposed framework, incorporating data from similar profiles could significantly improve the predictive performance of personalized models. Such methodology could produce more accurate predictions, enhancing the precision of personalized approaches.

\section*{Acknowledgements}
This study is part of the project ``New Science of Mental Disorders" (www.nsmd.eu), supported by the Dutch Research Council and the Dutch Ministry of Education, Culture and Science (NWO gravitation grant number 024.004.016).

\section*{Disclosure of Interest}
Authors have no conflict of interest to declare.

\bibliographystyle{splncs04}
\bibliography{Bibliography}

\begin{thebibliography}{10}
\providecommand{\url}[1]{\texttt{#1}}
\providecommand{\urlprefix}{URL }
\providecommand{\doi}[1]{https://doi.org/#1}

\bibitem{x-kmeans2}
Bandyapadhyay, S., Fomin, F.V., Golovach, P.A., Lochet, W., Purohit, N., Simonov, K.: How to find a good explanation for clustering? Artificial Intelligence p. 103948 (2023)

\bibitem{becker2016predict}
Becker, D., Bremer, V., Funk, B., Asselbergs, J., Riper, H., Ruwaard, J.: How to predict mood? delving into features of smartphone-based data  (2016)

\bibitem{timeshap}
Bento, J., Saleiro, P., Cruz, A.F., Figueiredo, M.A., Bizarro, P.: Timeshap: Explaining recurrent models through sequence perturbations. In: Proceedings of the 27th ACM SIGKDD Conference on Knowledge Discovery \& Data Mining. pp. 2565--2573 (2021)

\bibitem{time2feat}
Bonifati, A., Buono, F.D., Guerra, F., Tiano, D.: Time2feat: learning interpretable representations for multivariate time series clustering. Proceedings of the VLDB Endowment  \textbf{16}(2),  193--201 (2022)

\bibitem{cuturi2011fast}
Cuturi, M.: Fast global alignment kernels. In: Proceedings of the 28th international conference on machine learning (ICML-11). pp. 929--936 (2011)

\bibitem{cuturi2007kernel}
Cuturi, M., Vert, J.P., Birkenes, O., Matsui, T.: A kernel for time series based on global alignments. In: 2007 IEEE International Conference on Acoustics, Speech and Signal Processing-ICASSP'07. vol.~2, pp. II--413. IEEE (2007)

\bibitem{fried2017moving}
Fried, E.I., Cramer, A.O.: Moving forward: Challenges and directions for psychopathological network theory and methodology. Perspectives on Psychological Science  \textbf{12}(6),  999--1020 (2017)

\bibitem{hsieh2021explainable}
Hsieh, T.Y., Wang, S., Sun, Y., Honavar, V.: Explainable multivariate time series classification: a deep neural network which learns to attend to important variables as well as time intervals. In: Proceedings of the 14th ACM international conference on web search and data mining. pp. 607--615 (2021)

\bibitem{hsu2019multivariate}
Hsu, E.Y., Liu, C.L., Tseng, V.S.: Multivariate time series early classification with interpretability using deep learning and attention mechanism. In: Pacific-Asia Conference on Knowledge Discovery and Data Mining. pp. 541--553. Springer (2019)

\bibitem{xclusters}
Hwang, H., Whang, S.E.: Xclusters: explainability-first clustering. In: Proceedings of the AAAI Conference on Artificial Intelligence. vol.~37, pp. 7962--7970 (2023)

\bibitem{jacques2014functional}
Jacques, J., Preda, C.: Functional data clustering: a survey. Advances in Data Analysis and Classification  \textbf{8},  231--255 (2014)

\bibitem{shap}
Lundberg, S.M., Lee, S.I.: A unified approach to interpreting model predictions. Advances in neural information processing systems  \textbf{30} (2017)

\bibitem{martinez2023developing}
Mart{\'\i}nez, A.J., Lemmens, L., Fried, E.I., Roefs, A.: Developing a transdiagnostic ecological momentary assessment protocol for psychopathology.  (2023)

\bibitem{x-kmeans}
Moshkovitz, M., Dasgupta, S., Rashtchian, C., Frost, N.: Explainable k-means and k-medians clustering. In: International conference on machine learning. pp. 7055--7065. PMLR (2020)

\bibitem{ntekouli2022using}
Ntekouli, M., Spanakis, G., Waldorp, L., Roefs, A.: Using explainable boosting machine to compare idiographic and nomothetic approaches for ecological momentary assessment data. In: International symposium on intelligent data analysis. pp. 199--211. Springer (2022)

\bibitem{ntekouli2023evaluating}
Ntekouli, M., Spanakis, G., Waldorp, L., Roefs, A.: Evaluating multivariate time-series clustering using simulated ecological momentary assessment data. Machine Learning with Applications  \textbf{14},  100512 (2023)

\bibitem{ntekouli2023modelbased}
Ntekouli, M., Spanakis, G., Waldorp, L., Roefs, A.: Model-based clustering of individuals' ecological momentary assessment time-series data for improving forecasting performance (2023)

\bibitem{local}
Ozyegen, O., Prayogo, N., Cevik, M., Basar, A.: Interpretable time series clustering using local explanations. arXiv preprint arXiv:2208.01152  (2022)

\bibitem{lime}
Ribeiro, M.T., Singh, S., Guestrin, C.: Model-agnostic interpretability of machine learning. arXiv preprint arXiv:1606.05386  (2016)

\bibitem{roefs2022new}
Roefs, A., Fried, E.I., Kindt, M., Martijn, C., Elzinga, B., Evers, A.W., Wiers, R.W., Borsboom, D., Jansen, A.: A new science of mental disorders: Using personalised, transdiagnostic, dynamical systems to understand, model, diagnose and treat psychopathology. Behaviour Research and Therapy  \textbf{153},  104096 (2022)

\bibitem{ts_mule}
Schlegel, U., Vo, D.L., Keim, D.A., Seebacher, D.: Ts-mule: Local interpretable model-agnostic explanations for time series forecast models. In: Joint European Conference on Machine Learning and Knowledge Discovery in Databases. pp. 5--14. Springer (2021)

\bibitem{vskrlj2020feature}
{\v{S}}krlj, B., D{\v{z}}eroski, S., Lavra{\v{c}}, N., Petkovi{\v{c}}, M.: Feature importance estimation with self-attention networks. arXiv preprint arXiv:2002.04464  (2020)

\bibitem{integrated_grad}
Sundararajan, M., Taly, A., Yan, Q.: Axiomatic attribution for deep networks. In: International conference on machine learning. pp. 3319--3328. PMLR (2017)

\bibitem{torous2018smartphones}
Torous, J., Larsen, M.E., Depp, C., Cosco, T.D., Barnett, I., Nock, M.K., Firth, J.: Smartphones, sensors, and machine learning to advance real-time prediction and interventions for suicide prevention: a review of current progress and next steps. Current psychiatry reports  \textbf{20}, ~1--6 (2018)

\bibitem{vaswani2017attention}
Vaswani, A., Shazeer, N., Parmar, N., Uszkoreit, J., Jones, L., Gomez, A.N., Kaiser, {\L}., Polosukhin, I.: Attention is all you need. Advances in neural information processing systems  \textbf{30} (2017)

\bibitem{varmax}
Villani, M., Lockhart, J., Magazzeni, D.: Feature importance for time series data: Improving kernelshap. arXiv preprint arXiv:2210.02176  (2022)

\end{thebibliography}

\end{document}